\documentclass[cameraready]{Interspeech}
\title{Exploring the potential and limitations of Model Merging for Multi-Domain Adaptation in ASR
}

\author[affiliation={1,2}, orcid=0000-0002-9603-2423]{Carlos}{Carvalho}
\author[affiliation={1}, orcid=0000-0001-6614-3168]{Francisco}{Teixeira}
\author[affiliation={1}, orcid=0000-0003-2419-1377]{Thomas}{Rolland}
\author[affiliation={1,2}, orcid=0000-0003-2122-5148]{Alberto}{Abad}
\address{
    $^1$INESC-ID \& $^2$Instituto Superior Técnico, Universidade de Lisboa, Portugal 
}

\email{carlos.mf.carvalho@inesc-id.pt}


\def\ul#1 {\underline{#1} }
\usepackage{comment}
\usepackage{multirow}

\begin{document}

\maketitle

% the abstract here must exactly match the abstract entered into the paper submission system
\begin{abstract}
%\vspace{-0.1cm}
    % 1000 characters. ASCII characters only. No cita%tions.
Model merging is a scalable alternative to multi-task training that combines the capabilities of multiple specialised models into a single model. This is particularly attractive for large speech foundation models, which are typically adapted through domain-specific fine-tuning, resulting in multiple customised checkpoints, for which repeating full fine-tuning when new data becomes available is computationally prohibitive. In this work, we study model merging for multi-domain ASR and benchmark 11 merging algorithms for 10 European Portuguese domains, evaluating in-domain accuracy, robustness under distribution shift, as well as English and multilingual performance. We further propose BoostedTSV-M, a new merging algorithm based on TSV-M that mitigates rank collapse via singular-value boosting and improves numerical stability. Overall, our approach outperforms full fine-tuning on European Portuguese while preserving out-of-distribution generalisation in a single model.
%Model merging is a scalable alternative to multi-task training that aims to consolidate the capabilities of multiple specialised models into a single model. This is particularly interesting for large speech foundation models, which are typically adapted via domain-specific fine-tuning, yielding numerous specialised checkpoints that complicate maintenance and deployment at scale. Moreover, repeating full fine-tuning whenever new data or domains arrive is prohibitively costly. Therefore, we study parameter-space merging for multi-domain ASR and systematically evaluate eleven merging algorithms across ten European Portuguese domains, with regard to in-domain performance and robustness under distribution shift, as well as evaluation on English and multilingual test sets. We further propose BoostedTSV-M, which mitigates rank collapse through singular-value boosting and improves numerical stability relative to TSV-M. Overall, the proposed approach achieves the best in-domain results while preserving competitive out-of-domain generalisation within a single deployable model.
  
\end{abstract}

\vspace{-0.2cm}
\section{Introduction}
Large Speech Foundation Models (LSFMs) have recently become the dominant paradigm in Automatic Speech Recognition (ASR). This shift is driven by the availability of vast multilingual datasets and the massive computational resources required to scale Transformer-based architectures \cite{radford2023whisper,owsm4,canary, saon2025granitespeechopensourcespeechawarellms, Qwen3-ASR}. While these models exhibit outstanding in-distribution (ID) performance and out-of-distribution (OOD) generalisation \cite{radford2023whisper, Qwen3-ASR}, they are rarely ``one-size-fits-all." When labeled data for a specific target domain is accessible, fine-tuning remains the most effective strategy for bridging the gap to the target distribution~\cite{carvalho2025camoes}.
In practice, this adaptation is often done separately for each target domain. Over time, this approach yields a proliferation of domain-specific checkpoints -- in effect one fine-tuned model per domain. Such fragmentation complicates maintenance and deployment, as the system will have to identify the relevant domain and load the corresponding model for each request at inference time.
%Over time, this produces many specialised checkpoints -- i.e., one fine-tuned model per domain -- which are difficult to maintain and deploy, since the appropriate fine-tuned model must be selected and loaded depending on the task at hand. 
A straightforward and common alternative is to train a single model jointly on all available target datasets, but this can become impractical.
%A straightforward alternative is to train a single model jointly on all available target datasets, but this is often impractical.
Joint fine-tuning typically requires access to all previously used adaptation data, which may be no longer available, due to privacy or storage limitations. Moreover, incorporating a new target domain or dataset typically requires an additional large-scale fine-tuning run over the full collection of previously used and new datasets, with careful rebalancing of the training mixture to avoid performance degradation on previously supported domains, thereby further increasing computational and training time costs.  
%Moreover, adding new data can be expensive because it often forces another large fine-tuning run, where the training mix must be rebalanced to keep performance from dropping on previously supported domains.

Continual learning is a natural way to address this challenge, since it aims to update a model as new data or tasks become available without losing previously learned capabilities \cite{sadhu20_interspeech,wang2024comprehensivesurveycontinuallearning}. However, most methods are inherently sequential, requiring careful mechanisms such as replay buffers \cite{lopezpaz2022gradientepisodicmemorycontinual, shin2017continuallearningdeepgenerative}, regularisation \cite{li2017learningforgetting, Kirkpatrick_2017}, or dynamic architectures \cite{sadhu20_interspeech} to mitigate catastrophic forgetting. Furthermore, performance can depend strongly on the order in which domains are encountered \cite{tang2025mergingmodelsflyretraining}. Instead, we aim to explore \textbf{model merging} as an alternative approach: independently fine-tuned domain-specific models are combined into a single unified model, eliminating the need for sequential training, cached data, or additional parameters.

Model merging has been explored extensively in natural language processing (NLP) and computer vision (CV) \cite{ruan2025taskspecificmodelsunifiedsystems}, but has received comparatively little attention in ASR. Existing work covers low-resource languages~\cite{nagasawa2025}, disordered speech \cite{ducorroy2025robustfinetuningspeechrecognition}, children's ASR \cite{10887889, rolland2025groupawarepartialmodelmerging}, and speech synthesis \cite{murata2024attributeinterpolationmethodspeech}, as well as cross-lingual \cite{zhao2025lowranksparsemodelmerging} and cross-task settings \cite{sundar2024multimodalattentionmergingimproved}. 
%The most remarkable example exploring model merging for multi-domain ASR adaptation combines models fine-tuned on five distinct English domains \cite{10832275}, but only focuses on model averaging \cite{wortsman2022modelsoupsaveragingweights} and does not incorporate the original foundation model, leaving OOD generalisation unexplored. 
Notably, model merging for multi-domain ASR adaptation has previously been explored for English \cite{10832275}. However, this study was restricted to model averaging \cite{wortsman2022modelsoupsaveragingweights} and excluded the original foundation model, leaving OOD generalisation unexplored.% but only focusing on model averaging \cite{wortsman2022modelsoupsaveragingweights} and not incorporating the original foundation model, leaving OOD generalisation unexplored. 
%To our knowledge, no prior work has systematically compared a broader range of merging approaches for multi-domain ASR within a single language.
%To our knowledge, no prior work has systematically compared a broader set of merging approaches for multi-domain ASR within the same language. 

In this work, we go beyond prior studies by systematically evaluating
%benchmarking
eleven merging approaches across ten different European Portuguese (EP) domains \cite{carvalho2025camoes}. To support this comparison,
%benchmarking
we introduce \texttt{MergeWhisper}\footnote{\url{https://github.com/Miamoto/mergewhisper}}, an extension of \texttt{mergekit} \cite{goddard-etal-2024-arcees} that adds native Whisper support -- which the original toolkit lacks -- and incorporates all of the merging approaches available in the toolkit, together with additional methods explored in this work.
Moreover, we evaluate both ID performance for EP as well OOD generalisation across diverse conditions and language varieties, including African, Asian and Brazilian Portuguese \cite{carvalho2025camoes}. We also evaluate the resulting models on the Hugging Face Open ASR Leaderboard (OpenASR-HF) \cite{srivastav2025openasrleaderboardreproducible} and a subset of FLEURS \cite{conneau2022fleursfewshotlearningevaluation} derived from \cite{owsm4}. This evaluation is intended to assess if the model's original multilingual capabilities are preserved following the merging process.
Finally, we propose an improved version of the merging method that demonstrated the strongest performance in ID data (TSV-M~\cite{gargiulo2025tasksingularvectorsreducing}), called BoostedTSV-M, which (1) addresses rank collapse by boosting small singular values~\cite{skorobogat2025subspaceboostedmodelmerging}, and (2) improves algorithmic stability when merging closely related models by replacing the singular vector orthogonalization algorithm.

%Finally, we propose improvements to the strongest merging method for in-domain data we identified, TSV-M \cite{gargiulo2025tasksingularvectorsreducing}:
%, we further propose two improvements: 
%(1) a simple boosting of smaller singular values to prevent rank collapse \cite{skorobogat2025subspaceboostedmodelmerging}; and (2) an algorithmic stability improvement to TSV-M (and ISO-CTS) when merging closely related models. 

%, replacing the Procrustes orthogonalisation step with Newton-Schulz iterations, which yields more numerically stable solutions and allows the method to work for higher ranks.      
%Overall, the main contributions of this work can be summarized as follows:
%\begin{itemize}
%\item We benchmark eleven merging variants for multi-domain ASR and release \texttt{MergeWhisper}, a Whisper-compatible model merging toolkit derived from \texttt{mergekit}.
%\item We systematically evaluate the in-domain (ID) versus OOD trade-off across multiple European Portuguese domains and diverse OOD conditions, including Portuguese varieties, English, and multilingual FLEURS test sets.
%\item We propose \texttt{BoostTSV-M}, a novel merging approach achieving the best in-domain performance in our benchmark through singular value boosting and Newton-Schulz orthogonalisation.
%\end{itemize}
\vspace{-0.1cm}
\section{Merging methods}
\label{sec:merging_methods}
%In this section, we describe the merging methods explored in this work.
%We group existing methods into three categories based on the parameter space they operate on: parameter space merging , delta-space merging , and delta-subspace merging, which we detail next. 

%Let $\theta_0$ denote the pre-trained model parameters and $\theta_k$ the parameters after fine-tuning on domain $k$. We define the delta update, also called the task vector, as $\Delta_k = \theta_k - \theta_0$. For all merging methods we denote the final merged model as $\theta^\star$.

%CARLOS: removing method description: In this section, we describe the merging methods explored in this work. Methods that operate directly on the full model parameters $\theta_t$ are grouped under parameter-space merging. Methods that operate on the difference between fine-tuned and pre-trained parameters are grouped under delta-space merging. Finally, methods that decompose these differences into low-rank subspaces are grouped under delta-subspace merging; we describe this family in greater detail, as it is most closely related to our proposed approach.

%In this section, we outline the merging methods explored in this work.
 %For all merging methods we denote the final merged model as $\theta^\star$.

Model merging is a growing area of research that combines the weights of multiple fine-tuned variants -- typically of the same base model -- into a single model without retraining~\cite{wortsman2022model}. Existing approaches can be broadly grouped into three categories based on the parameter space they operate on: parameter-space merging (PS), $\tau$-space merging ($\tau$Spa), and $\tau$-subspace merging ($\tau$Sub).
Parameter-space methods operate directly on the full model parameters $\theta_t$ as a flattened vector, and include approaches such as Model Soups (SOUP)~\cite{wortsman2022model}, Model Stock~\cite{jang2024model}, Karcher mean~\cite{mancinelli2023computing}, and Multi-Spherical Linear Interpolation (SLERP)~\cite{goddard-etal-2024-arcees}.
$\tau$-space merging methods operate over the task-vector, i.e., the difference between fine-tuned and pre-trained parameters. Task Arithmetic (TA)~\cite{ilharco2022editing}, TrIm, Elect, and Sign (TIES)~\cite{yadav2023ties}, Parameter Competition Balancing (PCB)~\cite{du2024parameter}, and Select, Calculate and Erase (SCE)~\cite{wan2025fusechat} belong to this group. 
Finally, $\tau$-subspace methods operate over $\tau$ matrices in their original form (i.e., not flattened), and perform merging in low-rank subspaces. This family of methods represents the current state-of-the-art in CV and NLP, and includes approaches such as Task Singular Vectors Merging (TSV-M)~\cite{gargiulo2025tasksingularvectorsreducing}, Isotropic Merging in Common Subspace (ISO-C), and Isotropic Merging in Common and Task-Specific
Subspaces (ISO-CTS)~\cite{marczak2025no}. All of the merging methods described above are applied with the foundational model, except for SOUP, which we evaluate both with and without it. Also, all of these merging methods are supported by \texttt{MergeWhisper}.   
%
%Due to space constraints, 
We refer the reader to the original publications for full algorithmic details and default hyperparameters. %For continual learning, given the substantial methodological differences with traditional approaches, we consider a comparison with continual-learning approaches to be outside the scope of this work. %In the next section, we describe TSV-M in further detail, as it is the basis for our proposed method, BoostedTSV-M.

\section{BoostedTSV-M}
Let $\theta_0$ denote a pre-trained foundation model and $\theta_t$ the parameters obtained after fine-tuning $\theta_0$ on a task/domain $t$ (for $t=1,\dots,T$). 
We define the difference between the fine-tuned  update and the baseline, also called the task vector, as $\tau_t = \theta_t - \theta_0$.

\subsection{TSV-M}

TSV-M~\cite{gargiulo2025tasksingularvectorsreducing} exploits the low-rank structure of task vectors by decomposing each weight matrix $W_i$ within each $\tau_t$ using Singular Value Decomposition (SVD):
\begin{equation}
    U_{i, t}, \Sigma_{i, t}, {V^{\intercal}_{i, t}} = \mbox{SVD}(\tau_{i, t})
\end{equation}
Of the full rank, only a small fraction (e.g., $1/T$) of singular values and left and right singular vectors are retained, ensuring that only the dominant directions of each task vector are maintained.
%\begin{equation}
%    U_{k_{1:z}}, \Sigma_{k_{1:z}}, {V^{\intercal}_{k_{1:z}}}
%\end{equation}
%where $z$ is a percentage of the full rank.  
The retained singular vectors and singular values from all task vectors are then concatenated:
\begin{equation}
\label{eq:tsvm_concatenation}
\begin{split}
    \mathbf{U} = [U_1|U_2|...|U_T] \qquad \mathbf{V}^{\intercal} = [V_1|V_2|...|V_T]^{\intercal} \\
    \mathbf{\Sigma} = \mbox{block\_diag}{(\Sigma_1, \Sigma_2, ..., \Sigma_T)}
\end{split}
\end{equation}
Additionally, to mitigate task interference, the concatenated matrices of singular vectors are whitened, and therefore decorrelated, by solving the orthogonal Procrustes problem  \cite{gargiulo2025tasksingularvectorsreducing}, resulting in $\mathbf{U}_{orth}$ and $\mathbf{V}_{orth}$.
Finally, these three matrices are used to reconstruct the merged task vector matrix:
\begin{equation}
    \tau^{\text{TSV-M}} = \mathbf{U}_{orth} \mathbf{\Sigma} \mathbf{V}_{orth}^{\intercal}
\end{equation}
The final merged model is then obtained as $\theta^\star = \theta_0 + \lambda\,\tau^{\text{TSV-M}}$.

\subsection{BoostedTSV-M}

We enhance TSV-M with a singular-value boosting scheme inspired by prior work in CV~\cite{skorobogat2025subspaceboostedmodelmerging}. In our setting, the singular values of the task vector matrices decay sharply, so that after truncation and concatenation (eq. \ref{eq:tsvm_concatenation}) many components are suppressed, leading to rank collapse and loss of task-specific signals. To mitigate this, we boost the singular values $\Sigma_{i,t}$ of each weight matrix $W_i$ prior to eq. \ref{eq:tsvm_concatenation}.

%To mitigate this, for each SVD $\Sigma_{i,t}$ we choose the smallest index $s^*$ whose cumulative energy exceeds a threshold $\beta$,
%\begin{equation}
%s^* = \min\bigl\{\, s \in \{1,\dots,r\} : \tfrac{\sum_{j=1}^{s} \sigma_{i,t,j}}{\sum_{j=1}^{r} \sigma_{i,t,j} + \epsilon} \ge \beta \,\bigr\},
%\end{equation}
%and clamp all smaller singular values to $\sigma_{i,t,s^*}$:
%\begin{equation}
%\hat{\sigma}_{i,t,j} = \max(\sigma_{i,t,j},\, \sigma_{i,t,s^*}).
%\end{equation}
%TSV-M then proceeds unchanged, using $\hat{\Sigma}_{i,t}$ in place of $\Sigma_{i,t}$.

% ALBERTO - Vou comentar isto e propor uma alternativa
%For each $\Sigma_{i,t}$, we compute the cumulative sum of the singular values and normalise it by their total sum: 
%\begin{equation}
%c_{i,t}(s) = \frac{\sum_{j=1}^{s} \sigma_{i,t,v}}{\sum_{v=1}^{r} \sigma_{i,t,v} + \epsilon},
%\end{equation} where $\sigma_{i,t,v}$ denotes the $v$-th singular value of $\Sigma_{i,k}$, $r$ is the total number of singular values, $\epsilon$ is a small value that avoids division by 0 and $c_{i,t}(s) \in [0,1]$ captures the fraction of total energy explained by the first $s$ singular values. 
For each $\Sigma_{i,t}$, 
we define the function $c_{i,t}(s) \in [0,1]$ as the fraction of total energy explained by the first $s$ singular values:
\begin{equation}
c_{i,t}(s) = \frac{\sum_{j=1}^{s} \sigma_{i,t}^{(j)}}{\sum_{j=1}^{r} \sigma_{i,t}^{(j)} + \epsilon},
\end{equation} 
where $r$ is the total number of singular values  of $\Sigma_{i,t}$, $\sigma_{i,t}^{(j)}$ denotes the $j$-th singular value, and $\epsilon$ is a small value to avoid division by 0.
We then find the smallest index $s^*$ such that the cumulative energy $c_{i,t}(s)$ exceeds a threshold $\beta \in [0,1]$:
\begin{equation}
s^* = \min\bigl\{\, s \in \{1,\dots,r\} : c_{i,t}(s) \ge \beta \,\bigr\},
\end{equation} and boost all singular values below $\sigma_{i,t}^{(s^*)}$ by clamping them to that value:
\begin{equation}
\hat{\sigma}_{i,t}^{(j)} 
= \max( \sigma_{i,t}^{(j)}, \, \sigma_{i,t}^{(s^*)} ),
\qquad \forall\, j \in \{1, \dots, r\},
\end{equation}
effectively setting a lower bound on the singular values and preventing the suppression of task-specific information carried by smaller singular values in the final merged model. After this boosting operation, TSV-M proceeds as described previously, concatenating the singular values and vectors across tasks. % corresponding singular vectors across tasks.

\section{Experimental setup}

\subsection{Corpora}
\label{subsection:corpora}

In all of our experiments, we use a training set of ten European Portuguese (EP) corpora, corresponding to $\sim350$h of speech. Each of these datasets has a corresponding test partition that is used to evaluate ID performance. For EP OOD, we consider 
%four additional datasets; PT\_Children is evaluated in two age groups (3–6 and 7–10), yielding 
five additional OOD test partitions, totalling
%in Table~\ref{table:corpora}. 
%These OOD sets total
10.9h. The combined ID and OOD test set amounts to 46.2h, matching the CAMÕES benchmark test set~\cite{carvalho2025camoes}. Summary statistics for all corpora are provided in Table~\ref{table:corpora}; for additional details, we refer the reader to \cite{carvalho2025camoes}. 
In our experiments, we also explore the performance of the merged models in unseen Portuguese variants, namely African and Asian Portuguese (AAP) and Brazilian Portuguese (BP). % -- the most commonly spoken variety of the Portuguese language. 
For AAP, we use the AAP subset of Português Falado~\cite{bettencourt2000portugues} ($\sim$3.5h). For BP, we evaluate on the BP subset of Português Falado as well as the CORAA~\cite{candido2023coraa} and MuPe~\cite{leal2025mupe} test sets (total $\sim$46h). 
To assess whether merging preserves non-EP capabilities, we evaluate on two OOD benchmarks: OpenASR-HF for English\cite{srivastav2025openasrleaderboardreproducible}, 
%, (reporting the mean WER over its evaluation suite),
%(AMI\cite{kraaij2005ami}, Earnings22~\cite{del2022earnings},GigaSpeech~\cite{chen2021gigaspeech}, LibriSpeech test-clean/test-other~\cite{panayotov2015librispeech}, SPGISpeech~\cite{o2021spgispeech}, TED-LIUM~\cite{rousseau2012ted}, VoxPopuli~\cite{wang2021voxpopuli}),
and a subset of 21 languages of FLEURS~\cite{conneau2022fleursfewshotlearningevaluation, owsm4}. 
% (following \cite{owsm4}). %the following subset of languages: bel, ben, cat, deu, eng, fra, glg, hin, ind, ita, jpn, kor, nld, por, rus, spa, tha, tur, ukr, vie, and zho. 

\begin{table}[t]
\caption{European Portuguese (EP) corpora used in this work: in-domain (ID) training data and ID/out-of-domain(OOD) evaluation sets, with hours of speech and number of speakers per split. OOD corpora are used for evaluation only (train shown as “–”).} 
\centering
\resizebox{1.0\columnwidth}{!}{
\label{table:corpora}
\begin{tabular}{cl cc cc}
\toprule
& & \multicolumn{2}{c}{\bf Train} & \multicolumn{2}{c}{\bf Test} \\ \cmidrule{3-4} \cmidrule{5-6}
& \bf Corpus & \bf \#hours & \bf \#speakers & \bf \#hours & \bf \#speakers \\ %& \multicolumn{1}{c}{\bf Notes}   \\
\midrule
\multirow{9}{*}{\bf ID} 
& Alert (AL)~\cite{trancoso2003evaluation} & 45.6 & 1,356 & 6.6 & 175 \\% & Broadcast news data. \\
& BD-Publico (BDP) ~\cite{neto1997design}  & 21.8 & 100 & 2.0 & 10 \\ %& Read sentences extracted from an EP newspaper. \\
& Fala Bracarense (FP) ~\cite{falabracarense} & 66.1 & 75 & 6.1 & 8 \\ %&  Recorded in the city of Braga, collected between 2009-2014.\\
& Lectra (LE)~\cite{trancoso2008lectra} & 22.0 & 7 & 2.6 & 7 \\ %& University lectures. Speakers are shared among partitions. \\ 
& MLS\_extended (MLSE)~\cite{mls} & 54.8 & 12 & 1.0 & 10 \\ %& EP extension of MLS~\cite{mls}: automatically aligned audiobooks. \\
& MuAViC (MV)~\cite{muavic} & 19.2 & 100 & 0.4 & 2 \\ %& TEDx talks. \\
& PostPort (PP)~\cite{meinedo2010l2f} & 31.3 &  {\scriptsize $>$}247 & 3.9 & {\scriptsize $>$}30 \\ %& Debates and entertainment (a few  documentaries and information). \\
& PT\_Adults (PTA)~\cite{pt_adults} & 7.3 & 66 & 1.6 & 17 \\ %& Corresponds to YMA in~\cite{pt_adults}.\\
& PT\_Elderly (PTE)~\cite{pt_elderly} & 48.2 & 794 & 1.3 & 172 \\ %& Train/test speakers are aged between 60-75/76-100 years, 55 speakers in train have an unknown age $<$59. \\
& SpeechDat (SpD)~\cite{hagen2003hmm} & 30.2 & 3,349 & 9.7 & 604 \\ %& Telephone speech sampled at 8kHz, upsampled to 16kHz. \\ 
\midrule
\multirow{5}{*}{\bf OOD} & CommonVoice (CV)~\cite{commonvoice} & -- & -- & 1.8 & 42 \\ %& Speaker count estimated from the client ids provided in the corpus.\\
& PT\_Children 3--6 (PTC1)~\cite{pt_children} & -- & -- & 0.3 & 16 \\ %&  Corpus of child   speech. \\ 
& PT\_Children 7--10 (PTC2)~\cite{pt_children} & -- & -- & 1.7 & 36 \\ %&  Corpus of child   speech. \\ 

& PT Fundamental (PTF)~\cite{portuguesfundamental} &  -- & -- & 4.2 & 169 \\ %&  Low quality recordings of interviews collected in the 1970's. \\
& VoxCelebPT (VoxPT)~\cite{voxcelebpt} &  -- & -- & 2.9 & 13 \\ \cmidrule{2-6} %& Voices of Portuguese celebrities collected from YouTube. \\
& \multicolumn{1}{c}{\bf Total} & \bf 346.5 & \bf 6,106 & \bf  46.2 & \bf 1,311  \\ \bottomrule
\end{tabular}}
%}
\vspace{-0.3cm}
\end{table}

\subsection{Implementation}

HuggingFace~\cite{wolf2020transformers} is the primary framework that we use to implement and evaluate our work. The foundation model for all experiments is Whisper Large-v3 (WhisperLv3), chosen for its strong and robust performance in multilingual ASR \cite{radford2023whisper,Qwen3-ASR, carvalho2025camoes}. In this work, we considered two fine-tuning regimes: (i) full \textit{fine-tuning} (\textbf{Full-FT}), in which WhisperLv3 was fine-tuned jointly on the ten European Portuguese corpora mentioned above;
%in Section~\ref{subsection:corpora}—Alert (AL), BD-Publico (BDP), Fala Bracarense (FP), Lectra (LE), MLS\_extended (MLSE), MuAViC (MV), PostPort (PP), PT\_Adults (PTA), PT\_eldery (PTE), and SpeechDat (SpD)
and (ii) \textit{individual fine-tuning} (\textbf{ID-FT}), in which WhisperLv3 was fine-tuned separately on each corpus. For Full-FT, we trained for 10 epochs using a per-device batch size of 64 with gradient accumulation of 4 (effective batch size 256) and a learning rate of $1\times10^{-5}$. For ID-FT, we trained each corpus-specific model for 3 epochs using a per-device batch size of 4 with gradient accumulation of 64 (effective batch size 256) and a learning rate of $2\times10^{-6}$ to ensure stable optimisation and convergence; higher learning rates led to unstable training and, in some cases, prevented the model from converging. In both settings, we used a cosine learning rate schedule with a warm-up phase covering 20\% of the total training steps. Utterances longer than 30 seconds are excluded during training and fine-tuning, to match Whisper’s common training recipe. Finally, Full-FT was conducted on a single NVIDIA A100 (80GB) GPU, whereas ID-FT was conducted on a single NVIDIA RTX A6000 (48GB) GPU.   

Inference with WhisperLv3 is known to produce hallucinations \cite{baranski2025investigation}. We mitigate this by applying VAD-based segmentation prior to decoding using WhisperX \cite{whisperx}; we adopt this configuration (WhisperLv3-X) for all subsequent experiments, using a beam size of 5. All models use consistent text normalisation during evaluation. For the English benchmark, we apply Whisper’s standard English normaliser; for EP, we use a Portuguese-specific normaliser derived from Whisper’s baseline normalisation procedure. For FLEURS, we report results without additional normalisation. Performance is reported in terms of word error rate (WER) and character error rate (CER) for languages without explicit word boundaries.
\vspace{-0.2cm}
\subsection{Model merging}
\vspace{-0.1cm}
\label{subsection:mm}
For all merging methods mentioned in Section~\ref{sec:merging_methods}, we use the default recommended hyperparameters from the original papers. % to allow for a fair comparison between the different methods. 
%In addition, preliminary experiments showed that the default configurations most often performed the best.
% unless stated otherwise. 
%For TSV-M (our strongest method on EP ID in preliminary experiments), we tuned the truncation fraction and found that retaining 10\% of the rank (128 singular values) performed best. 
For BoostedTSV-M, we swept $\beta \in [0.1,1.0]$ in preliminary experiments and found $\beta=0.3$ as the best performing configuration (cf. Figure~\ref{fig:beta}).
% on the Alert dataset development partition~\cite{alert}. !!!!!
%
The orthogonal Procrustes solution used by TSV-M and ISO-CTS methods was found to be numerically unstable in our setting when retaining more than 50\% rank for TSV-M and also prevented ISO-CTS merging from converging. 
%We hypothesize that this is due to the high similarity of the task vector matrices, which creates issues when running SVD over the stacked singular vectors.
We therefore replaced this step with Newton–Schulz orthogonalisation for both methods, as well as our own proposed method (NVIDIA's implementation~\cite{nemo_orthogonalized_optimizers})
%\footnote{\url{https://docs.nvidia.com/nemo/emerging-optimizers/latest/apidocs/orthogonalized-optimizers.html}}),
using 5 iterations and a \texttt{quintic} coefficient schedule; we view this stable substitution as a practical contribution that enables high rank-percentage retention.

\begin{table}[t]
\setlength{\tabcolsep}{4pt}
\caption{ASR error rates (\%) on in-domain EP (ID) and out-of-distribution evaluation sets (EP, AAP and BP), as well as OpenASR-HF and FLEURS. We report WER for all datasets except FLEURS, where we report the macro-average error rate across the selected languages (WER for languages with explicit word boundaries; CER otherwise).}
\centering
\scalebox{0.6}{
\label{tab:merged-results}
\begin{tabular}{c|c|c|c|c|c|c|c|c}
\toprule
\bf Group & \bf System & \bf EP ID & \bf EP OOD & \bf EP Full Avg. & \bf AAP & \bf BP & \bf OpenASR-HF & \bf Fleurs \\
\toprule
 & zero-shot & 15.62 & 25.21 & 18.82 & 29.00 & 22.17 & 7.17 & 12.06 \\
 & Full-FT & \bf 8.54 & 17.65 & \underline{11.58} & 23.96 & 27.00 & 8.83 & 16.25 \\
\midrule
\multirow{5}{*}{\bf PS} 
 & SOUP w/o pt. & 10.15 & 17.60 & 12.63 & 23.04 & 21.43 & 7.14 & 11.84 \\
 & SOUP & 11.25 & 17.97 & 13.65 & 23.88 & 21.55 & 7.14 & \underline{11.68} \\
 & Model Stock & 15.02 & 24.49 & 18.18 & 28.45 & 22.27 & 7.20 & \bf 11.46 \\
 & Karcher Mean & 10.23 & 17.63 & 12.70 & 22.88 & \bf 20.99 & \underline{7.12} & 11.79 \\
 & Multi-SLERP & 10.35 & 17.83 & 12.85 & 23.18 & 21.17 & 7.13 & 11.87 \\
\midrule
\multirow{4}{*}{\bf $\tau$Spa}
 & TA  & 10.23 & 17.63 & 12.70 & 22.90 & \underline{21.01} & \bf 7.12 & 11.78 \\
 & TIES & 10.45 & 17.76 & 12.89 & 23.78 & 27.62 & 8.77 & 19.77 \\
 & PCB & 9.79 & 16.94 & 12.17 & 22.19 & 23.02 & 7.17 & 12.99 \\
 & SCE & 9.76 & 16.73 & 12.08 & 21.86 & 23.32 & 7.18 & 13.87 \\
\midrule
\multirow{6}{*}{\bf $\tau$Sub}
 & ISO-C & 11.44 & 19.92 & 14.26 & 25.65 & 21.61 & 7.83 & 13.31 \\
 & ISO-CTS w/ NS & 12.59 & 20.99 & 15.39 & 26.20 & 21.75 & 7.49 & 12.30 \\
 & TSV-M w/ OP & 9.40 & 16.18 & 11.66 & \bf 21.53 & 22.07 & 7.31 & 13.46 \\
 & TSV-M w/ NS & 9.41 & \bf 16.07 & 11.63 & 21.61 & 22.12 & 7.24 & 13.46 \\
\cmidrule{2-9}
 & BoostedTSV-M & \underline{9.27} & \underline{16.11} & \bf 11.55 & \underline{21.58} & 24.98 & 7.60 & 14.43 \\
\bottomrule
\end{tabular}}
\vspace{-0.3cm}
\end{table}

\begin{table*}[t]
\caption{WER (\%) per EP corpus on in-domain (ID) and out-of-distribution (OOD) test sets. ``Avg.'' is the mean across corpora within each split. For ID-FT, we report the score of the model fine-tuned on each corresponding ID corpus (one model per domain), which serves as an upper bound for what a single merged model could achieve on ID performance.}
\label{tab:mann-results}
\centering
\resizebox{0.9\textwidth}{!}{
\begin{tabular}{l|cccccccccc|c|ccccc|c}
\toprule
& \multicolumn{10}{c}{\bf ID} & & \multicolumn{5}{c}{\bf OOD} & \\ \cmidrule{2-18}
\multicolumn{1}{c|}{\bf System} & AL & BDP & FB & LE & MLSE & MV & PP & PTA & PTE & SpD & Avg. & CV & PTC1 & PTC2 & PTF & VoxPT & Avg. \\
\toprule
zero-shot & 8.21 & 4.52 & 29.55 & 20.53 & \ul 9.21 & 12.72 & 18.04 & 18.94 & 26.7 & 7.76 & 15.62 & 9.98 & 42.25 & 12.15 & 49.12 & 12.56 & 25.21 \\
Full-FT & \ul 5.00 & \bf 1.71 & \bf 16.70 & \ul 9.46 & 9.32 & 13.91 & 14.91 & \bf 2.33 & \bf 7.67 & \bf 4.40 & \ul 8.54 & 10.35 & 25.26 & 4.40 & 39.27 & 8.95 & 17.65 \\
ID-FT & \bf 4.85 & \ul 2.42 & \ul 15.61 & \bf 8.74 & \bf 8.94 & \bf 11.55 & 15.48 & 2.95 & \ul 8.16 & \ul 5.15 & \bf 8.39 & -- & -- & -- & -- & -- & -- \\
\midrule
%SOUP w/o pt. & 6.37 & 3.83 & 20.75 & 11.94 & 9.51 & \ul 12.52 & 14.64 & 3.51 & 11.66 & 6.73 & 10.15 & \bf 9.83 & 24.74 & 4.84 & 39.65 & 8.94 & 17.60 \\
%SCE & 5.84 & 3.47 & 19.34 & 10.21 & 9.25 & 15.68 & \ul 14.11 & 3.22 & 10.39 & 6.04 & 9.76 & 10.05 & 22.65 & 4.41 & 38.12 & 8.41 & 16.73 \\
TSV-M w/ NS & 5.36 & 3.01 & 18.84 & 10.02 & 9.37 & 14.97 & \ul 14.11 & 2.97 & 9.76 & 5.65 & 9.41 & \ul 9.85 & \bf 20.64 & \bf 3.98 & \bf 37.61 & \bf 8.29 & \bf 16.07 \\ \midrule
BoostedTSV-M & 5.17 & 2.58 & 18.78 & 9.88 & 9.41 & 15.41 & \bf 13.93 & \ul 2.71 & 9.42 & 5.42 & 9.27 & 9.97 & \ul 20.73 & \ul 4.00 & \ul 37.46 & \ul 8.4 & \ul 16.11 \\
\bottomrule
\end{tabular}}
\vspace{-0.1cm}
\end{table*}

%\begin{table}[t]
%\setlength{\tabcolsep}{6pt}
%\caption{ASR error rates (\%) on FLEURS.}
%\label{tab:fleurs-results}
%\centering
%\resizebox{0.4\textwidth}{!}{
%\begin{tabular}{c|c|c|c|c|c}
%\toprule
%\bf Lang. & \bf Metric ↓ &\bf WhisperLv3-X & \bf Model Stock & \bf TA & \bf ISO-CTS \\
%\toprule
%bel & WER & 44,59 &\bf42,62&43,33& 42,83 \\
%ben & WER & 53,11 & 53,14&\bf51,68& 54,28 \\
%cat & WER & 6,48 &6,22&6,71& \bf 6,18 \\
%deu & WER & 6,25 &5,83&6,1& \bf 5,79 \\
%eng & WER & 5,92 & \bf5,67 &5,69& \bf 5,67 \\
%fra & WER & 7,02 &6,79&\bf 6,63& 6,82 \\
%glg & WER & 13,72 & 12,75&\bf 12,51& 12,8 \\
%hin & WER & 18,16 & 18,22&\bf18,01& 18,11 \\
%ind & WER & 7,43 &\bf 6,59&6,83&  6,72 \\
%ita & WER & 4,39 & \bf 4,24 &4,31& 4,3 \\
%jpn & CER & 7,58 &6,42&8,57& \bf 6,29 \\
%kor & CER & 2,81 &2,21&2,42& \bf 2,15 \\
%nld & WER & 7,12 &6,92&7,03& \bf 6,86 \\
%por & WER & 5,63 &5,39&7,53& \bf 5,28 \\
%rus & WER & \bf 6,16 &6,31&6,29& 6,25 \\
%spa & WER & 4,22 & 4,12 & 4,2& \bf 4,08 \\
%tha & CER & 14,58 &\bf 11,85&13,06& 11,99 \\
%tur & WER & 8,5 &7,61&7,71& \bf 7,53 \\
%ukr & WER & 7,69 & 7,46&\bf7,43&7,5 \\
%vie & WER & 9,6 & 8,7&\bf8,56& 8,91 \\
%zho & CER & 12,39 &11,54&12,69& \bf 11,19 \\
%\bottomrule
%\end{tabular}}
%\end{table}

\section{Results}

\subsection{Comparative analysis of merging methods}
%Baseline analysis
Table~\ref{tab:merged-results} presents our main results, including ID and OOD zero-shot and Full-FT performance, together with the model merging methods under evaluation, which are grouped according to the three paradigms introduced in Section~\ref{sec:merging_methods}: \emph{parameter-space} (PS), \emph{$\tau$-space} ($\tau$Spa), and \emph{$\tau$-subspace} ($\tau$Sub) merging.

%From Table~\ref{tab:merged-results},
%When comparing zero-shot and Full-FT, 
As expected, Full-FT 
%on the EP corpora 
substantially improves EP performance over the zero-shot model, reducing WER from 15.62\% to 8.54\% on the ID test sets and from 25.21\% to 17.65\% on EP OOD. A similar trend is observed for AAP, where WER decreases from 29.00\% to 23.96\%. 
In contrast, Full-FT worsens performance on all non-EP OOD data except AAP, indicating that specialising the model to EP can harm multilingual robustness, and potentially cause catastrophic forgetting. On the other hand, the fact that AAP improves under EP fine-tuning, while other OOD data (BP, OpenASR-HF and Fleurs) degrade, suggests that AAP is distributionally closer to the EP data.
%\subsection{Comparison of merging approaches}

%We group the methods in Table~\ref{tab:merged-results} according to the three paradigms introduced in Section~\ref{sec:merging_methods}: \emph{parameter-space}, \emph{delta-space}, and \emph{delta-subspace} merging. 

%As mentioned in the introduction, our goal with this work is to evaluate the potential of merging approaches as an alternative to Full-FT.
When evaluating merging approaches as alternatives to Full-FT, the $\tau$Spa and $\tau$Sub families contain the strongest merging algorithms for EP.
%Regarding the potential of merging approaches as an alternative to Full-FT, among the different merging paradigms, the $\tau$Sub methods are the ones that achieve better performance. %In this regard, among the different merging paradigms, EP performance is the strongest when using $\tau$Sub methods. 
Among these, TSV-M yields the best EP OOD performance as well as the best AAP result overall -- this being consistent with the claim that AAP is distributionally closer to the EP fine-tuning data than BP. % and the broader multilingual benchmarks. %, as mentioned above.
As an additional note, as stated in Section~\ref{subsection:mm}, we replaced the orthogonal Procrustes (OP) step with Newton--Schulz (NS) orthogonalisation in TSV-M and ISO-CTS; empirically, this substitution yields comparable performance across benchmarks when comparing the two versions of TSV-M, while improving numerical stability, and allowing us to apply ISO-CTS merging. %, and we adopt it in subsequent experiments.

%Finally, 
Our proposed modification of TSV-M, BoostedTSV-M achieves the best EP Full Avg. (11.55\% WER), slightly improving over Full-FT (11.58\% WER) with statistical 
significance ($p < 0.001$, obtained through the MAPSSWE test~\cite{gillick1989}). This improvement for EP comes at the cost of degrading non-EP OOD with respect to TSV-M. %; and, more generally
%This behaviour is consistent with our design choice: 
%boosting increases the contribution of task-specific components in the merged update, which can improve target-domain accuracy at the expense of some OOD robustness.

%Although they do not match the DSub variants on ID datasets, 
PS methods generally yield the best non-EP OOD performance, though they are not able to match the $\tau$Sub variants on ID data and EP OOD.
%While not as strong for ID datasets as DSub methods, 
%Overall, 
%PS methods tend to provide the strongest OOD performance on BP and FLEURS. % among the considered approaches. 
In particular, Karcher mean achieves the best BP result, improving 1.2\%
%22.17\% to 20.99\% 
absolute WER relative to the WhisperLv3-X baseline, while Model Stock attains the lowest average FLEURS error rate (11.46\%), a 0.60 WER improvement over WhisperLv3-X and the best result among all methods. 
Finally, $\tau$Spa methods slightly outperform PS methods in EP evaluation, but starkly degrade non-EP OOD (except for AAP).  

For OpenASR-HF, most merging methods 
%across all three paradigms 
remain close to the zero-shot baseline and in several cases even slightly improve it (e.g., Karcher mean, Multi-SLERP, and TA), whereas Full-FT incurs a clear degradation, which is consistent with catastrophic forgetting; notably, TIES is the only merging method whose English performance starkly degrades when compared to the baseline, with a very similar performance to Full-FT.

%In the experiments presented above, methods that emphasise task-specific components, namely DSub merging methods like TSV-M and our proposed BoostedTSV-M, improve EP performance on both ID and EP OOD, but often reduce robustness on non-target evaluations (BP, OpenASR-HF, and FLEURS).
%
%Conversely, more conservative merging strategies, like SOUP or Model Stock, better preserve -- and sometimes improve -- these OOD and multilingual benchmarks relative to WhisperLv3-X, but typically lag behind the EP-specialised Full-FT model on EP ID accuracy.
Overall, these experiments show that, while a relevant alternative to Full-FT, merging exhibits a clear trade-off between target-domain specialisation and cross-lingual robustness.

\begin{figure}[ht!]
  \centering
\includegraphics[width=0.40\textwidth]
{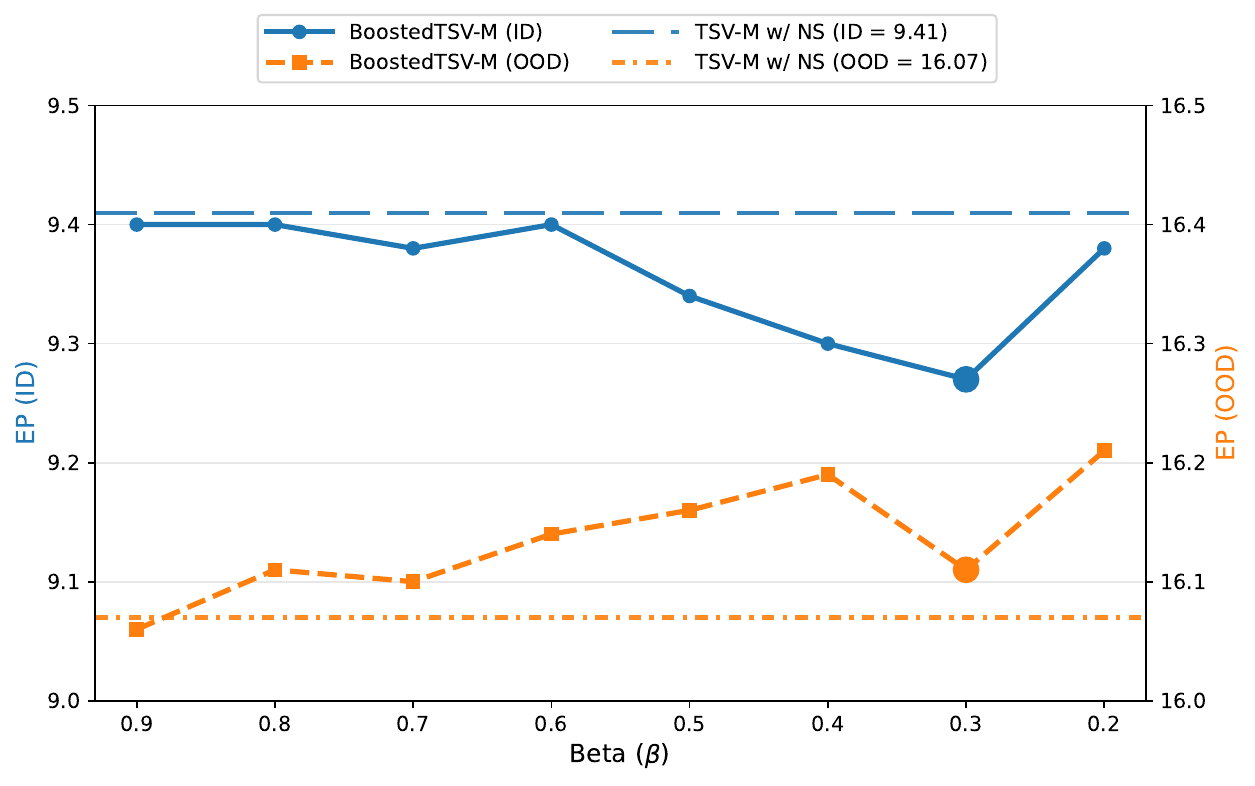}
  \caption{ID/OOD EP WER (\%) for BoostedTSV-M as a function of $\beta$. TSV-M w/ NS performance is shown as horizontal lines. 
  }
 \label{fig:beta}
 \vspace{-0.3cm}
\end{figure}

\subsection{Comparison of BoostedTSV-M and TSV-M}
We additionally compare the performance of TSV-M w/ NS and BoostedTSV-M for both ID and OOD (EP and non-EP) data. 
Table~\ref{tab:mann-results}, presents per-corpus WER on the EP test sets, enabling a fine-grained assessment of the impact of these methods in same-language performance (EP).
We further report the score of the model fine-tuned on each corresponding corpus, ID-FT, which serves as an upper bound for model merging and Full-FT.

%-- using the same corpus abbreviations as in Table~\ref{table:corpora} -- 
%for a more fine-grained analysis of same-language performance.
%For brevity, we only include the two best-performing merging methods for EP, namely TSV-M w/ NS and BoostedTSV-M. 

On the ID sets, BoostedTSV-M improves over TSV-M w/ NS on most corpora, with the exception of MV and MLSE. Notably, it achieves the second best result on PT-Adults (PTA), and the lowest WER on PostPort (PP), reaching 13.93\% and outperforming both Full-FT and the corresponding ID-FT model. On the EP OOD sets, TSV-M w/ NS and BoostedTSV-M further improve over Full-FT, reducing the average EP OOD WER by 1.58 and 1.54 points, respectively, with both merging methods outperforming the zero-shot and Full-FT baselines on all EP OOD corpora. 
The fact that TSV-M w/ NS outperforms BoostedTSV-M on all EP OOD corpora, and the opposite happening for most ID corpora, strengthens our claim that boosting trades-off target-domain performance for EP OOD performance.
Moreover, while neither method is able to achieve the performance of Full-FT or ID-FT on ID data, we argue that the method's ability to preserve the capabilities of the base foundation model compensates the ID performance gap.

On the other hand, from Table \ref{tab:merged-results}, we observe that performance on AAP for BoostedTSV-M remains close to TSV-M, but we observe degradation on BP, OpenASR-HF, and FLEURS when compared to TSV-M. 

To better understand the effect of boosting, we performed an ablation of the hyperparameter $\beta$ (Figure~\ref{fig:beta}) as mentioned in Section~\ref{subsection:mm}. ID performance improves as $\beta$ decreases, with diminishing returns beyond $\beta \approx 0.6$, while EP OOD performance generally degrades as $\beta$ decreases. 
These results reflect a fundamental trade-off that is consistent with our design choice: smaller values of $\beta$ amplify task-specific singular values at the expense of shared structure that supports transfer to OOD. 

\vspace{-0.2cm}
\subsection{OOD cross-lingual performance}
\vspace{-0.1cm}
%Table~\ref{tab:fleurs-results} reports results on the FLEURS language subset described in Section~\ref{subsection:corpora}.
The last column of Table~\ref{tab:merged-results} reports the average results obtained for 21 languages of the FLEURS dataset (mentioned in Section~\ref{subsection:corpora}).
We observe that a large number of merged models actually outperform
%, with the exception of Russian, 
%the best mean score 
%for each language 
%is achieved by one of the merged models rather%
%than
the WhisperLv3-X baseline. This indicates that through model merging it is possible to obtain gains even for languages that are not related to the fine-tuning data. We hypothesise that these improvements stem from shared acoustic characteristics (e.g., channel conditions, speaking styles, or noise profiles) captured by the diverse adaptation domains. More broadly, this suggests that merging domain-specialised models trained on a diverse set of acoustic conditions can improve robustness and transfer knowledge to languages that are otherwise unrelated, matching what has been previously reported in the literature~\cite{nagasawa2025}.

%\subsection{Discussion}
%In the experiments presented above, methods that emphasise task-specific components, namely DSub merging methods like TSV-M and our proposed BoostedTSV-M, improve EP performance on both ID and EP OOD, but often reduce robustness on non-target evaluations (BP, OpenASR-HF, and FLEURS).
%
%Conversely, more conservative merging strategies, like SOUP or Model Stock, better preserve -- and sometimes improve -- these OOD and multilingual benchmarks relative to WhisperLv3-X, but typically lag behind the EP-specialised Full-FT model on EP ID accuracy. Another such case is ISO-CTS which presents a comparatively poor ID performance, but the second best performance for multilingual OOD tasks. This is likely due to the fact that ISO-CTS is designed to emphasize the common subspace of the merged tasks, instead of task specific contributions, leading to an improvement of the already existing characteristics in the base model.
%
%

%Model merging actually solves these issues: many merging approaches preserve -- and in several cases improve -- BP, OpenASR-HF, and FLEURS performances relative to the zero-shot baseline, while remaining comparatively close to Full-FT on EP. Nevertheless, a subset of methods still exhibits degradation in specific OOD benchmarks, highlighting an inherent trade-off between ID specialisation and broader robustness. 
\vspace{-0.1cm}
\section{Conclusion}
\vspace{-0.1cm}
This work studied model merging for multi-domain ASR, benchmarking eleven merging algorithms for WhisperLv3-X adapted to ten European Portuguese datasets, evaluating both in-domain accuracy and robustness under domain and language shift, including other Portuguese varieties, OpenASR-HF (English), and FLEURS (multilingual). 
It also introduced \texttt{MergeWhisper}, a Whisper-compatible merging toolkit that implements the evaluated methods to support future ASR research. 
Results showed that while joint fine-tuning yields the best EP accuracy, it can substantially degrade performance on other Portuguese varieties and on English/multilingual benchmarks. 
In contrast, most merging approaches preserve -- and in several cases improve -- OOD performance, while narrowing the EP performance gap relative to joint fine-tuning; however, there exists a clear trade-off between EP specialisation and cross-domain/multilingual robustness. 
Building on TSV-M, this work further proposed BoostedTSV-M, which mitigates rank collapse via singular-value boosting, allowing it to achieve the strongest EP performance of all model merging methods under consideration.
% and  the strongest method for EP ID/OOD among the merging approaches, we propose BoostedTSV-M, . 
Overall, the findings of this work show that model merging provides a practical way to produce a single deployable model that achieves strong EP accuracy and improved generalization across unseen domains and languages.

\section{Acknowledgements}

Work supported by national funds through Fundação para a Ciência e a Tecnologia, I.P. (FCT) under projects UID/50021/2025 (DOI: https://doi.org/10.54499/UID/50021/2025),  UID/PRR/50021/2025 (DOI: https://doi.org/10.54499/UID/PRR/50021/2025) and 2022/12328/BD, as well as by the Portuguese Recovery and Resilience Plan (RRP) through project C644865762-00000008 (Accelerat.AI) and contract RE-C05-i08 (AMALIA). 

%Work supported by Portuguese national funds through Fundação para a Ciência e a Tecnologia (FCT), with references UIDB/50021/2020 and 2022/12328/BD, as well as by the Portuguese Recovery and Resilience Plan (RRP) through project C644865762-00000008 (Accelerat.AI).

\section{Generative AI Use Disclosure}
Generative AI tools have been used to help revise and refine the
manuscript and code. Authors take full responsibility for the content.

\bibliographystyle{IEEEtran}
\bibliography{mybib}
\clearpage

\end{document}